**PaveCap: The First Multimodal Framework for Comprehensive Pavement Condition Assessment with Dense Captioning and PCI Estimation.**

**Blessing Agyei Kyem**

PhD Student
North Dakota State University
Department of Civil, Construction and Environmental Engineering
Email: blessing.agyeikyem@ndsu.edu

**Eugene Kofi Okrah Denteh**

Research Assistant
Kwame Nkrumah University of Science and Technology
Department of Civil Engineering
Email: ekdenteh@st.knust.edu.gh

**Joshua Kofi Asamoah**

PhD Student
North Dakota State University
Department of Civil, Construction and Environmental Engineering
Email: joshua.asamoah@ndsu.edu

**Armstrong Aboah, Corresponding Author**

Assistant Professor
North Dakota State University
Department of Civil, Construction and Environmental Engineering
Email: armstrong.aboah@ndsu.edu

**ABSTRACT**

This research introduces the first multimodal approach for pavement condition assessment, providing both quantitative Pavement Condition Index (PCI) predictions and qualitative descriptions. We introduce PaveCap, a novel framework for automated pavement condition assessment. The framework consists of two main parts: a Single-Shot PCI Estimation Network and a Dense Captioning Network. The PCI Estimation Network uses YOLOv8 for object detection, the Segment Anything Model (SAM) for zero-shot segmentation, and a four-layer convolutional neural network to predict PCI. The Dense Captioning Network uses a YOLOv8 backbone, a Transformer encoder-decoder architecture, and a convolutional feed-forward module to generate detailed descriptions of pavement conditions. To train and evaluate these networks, we developed a pavement dataset with bounding box annotations, textual annotations, and PCI values. The results of our PCI Estimation Network showed a strong positive correlation (0.70) between predicted and actual PCIs, demonstrating its effectiveness in automating condition assessment. Also, the Dense Captioning Network produced accurate pavement condition descriptions, evidenced by high BLEU (0.7445), GLEU (0.5893), and METEOR (0.7252) scores. Additionally, the dense captioning model handled complex scenarios well, even correcting some errors in the ground truth data. The framework developed here can greatly improve infrastructure management and decision-making in pavement maintenance.

**INTRODUCTION**

Pavement condition assessment is crucial for evaluating the service conditions of road infrastructure. This process facilitates the early detection of structural issues, playing a pivotal role in maintaining serviceability and ensuring pavement longevity. Traditionally, experienced engineers conduct visual inspections to assess pavement conditions, focusing primarily on pavement distresses found on the pavement's surface. While this approach provides a detailed assessment of the pavement condition, this method suffers from inefficiency and subjectivity due to the variability in inspectors' experience, highlighting the need for an automated and consistent assessment process.

In recent years, the advent of digital image-based pavement assessment has accelerated the automated assessment process, thereby mitigating subjectivity in analytical processes traditionally performed by engineers. Early research in this area employed conventional image processing techniques, such as threshold segmentation, edge detection, and morphological operations, to identify pavement distress patterns *(1)*. Despite their foundational significance, these methods faced limitations in addressing the complexity and variability of real-world pavement conditions, necessitating more advanced approaches *(2)*. More recently, the emergence of machine learning brought significant improvements, mainly through feature engineering and the use of support vector machines for distress classification *(1)*. Despite these advancements, a significant paradigm shift occurred with the introduction of deep learning, which greatly enhanced the performance and robustness of computer vision algorithms. Today, deep learning-based object classification algorithms are extensively used for categorizing pavement distress, while object detection and segmentation algorithms accurately identify damaged areas *(3, 4)*.

Despite the numerous studies demonstrating the efficacy of deep learning models in identifying pavement distress, a significant challenge persists: pavement condition assessment remains a multimodal problem, but it is often oversimplified as a unimodal problem. That is, current approaches predominantly focus on numerical results derived from image classification, detection, or segmentation, which fail to fully capture the nuanced understanding necessary for effective communication and decision-making *(5)*. A highly abstract natural language description can more comprehensively summarize and convey the pavement assessment process and results to human stakeholders *(6)*. This gap in current research underscores the urgent need for innovative methodologies that incorporate natural language as a modality in assessing pavement conditions. Therefore, this study seeks to pioneer the use of vision-language models for dense pavement captioning, transforming the way pavement conditions are assessed and reported.

The emergence of dense captioning models (DCM) has enabled the training of joint embeddings of images and text, integrating text—a natural form of human understanding—into the analysis process. DCMs have demonstrated significant success in other domains by providing accurate assessments and generating human-understandable explanations for their predictions. This study seeks to explore the application of DCMs in pavement condition assessment, aiming to develop a model that can process pavement images and produce a natural language description of the visual representation. Our proposed model would also provide justification for the captioning through bounding box grounding and ultimately offer a PCI value for the given image.

The main goal of this research is to leverage the capabilities of Dense Captioning Models (DCMs) to transform the assessment of pavement conditions. The study presents two innovative, end-to-end operational frameworks. The first framework involves the development of a single-shot Pavement Condition Index (PCI) estimation architecture that is trained to identify various types of pavement distress and directly estimate the PCI from pavement images. The second framework is a network designed to process pavement images and generate detailed natural language descriptions, incorporating bounding box grounding for more precise localization. To tailor the DCM for this specific task, we initially prepared a dataset comprising pavement images paired with manually annotated textual descriptions detailing the pavement condition. These descriptions included the types of distresses present and absent in the pavement images and the overall PCI value. Additionally, we annotated bounding boxes to provide visual grounding information, guiding the model in recognizing the precise locations of the various distresses within the images. The model was then fine-tuned using this annotated dataset. To evaluate the efficacy of our fine-tuned model, we tested it on new pavement images, assessing its ability to generate accurate captions regarding pavement conditions. Furthermore, we compared the performance of our fine-tuned model against the original base model, which had not undergone fine-tuning with our dataset. To this end, the study makes the following contributions;

1. Developed a novel single-shot PCI estimation architecture capable of detecting various distress types and estimating PCI directly from pavement images. The model offers a streamlined approach to pavement condition assessment, significantly reducing the time and effort required compared to current deep learning methods.
2. Introduced the first pavement dense captioning model, which processes pavement images and produces detailed natural language descriptions for human understanding. By combining visual and textual data, the model enhances the interpretability of pavement condition assessments, enabling stakeholders to obtain comprehensive and understandable reports.
3. Created and made publicly available a comprehensive multimodal dataset for pavement condition assessment. This dataset includes pavement images paired with manually annotated textual descriptions of pavement conditions, bounding box annotations for various distress types, and corresponding PCI values. By making this dataset accessible to the research community, we aim to facilitate further advancements in automated pavement assessment techniques (https://github.com/Blessing988/PaveCap/tree/main)

By integrating dense natural language captioning into pavement condition assessment, both the accuracy and explainability of automated systems can be significantly enhanced. This integration would provide precise assessments and render the results more interpretable to a broader audience, including non-experts. This proposed system would improve transparency and trust, facilitating better communication between engineers, policymakers, and the public. The potential impact of this research is substantial. By offering natural language descriptions and visual justifications for detected pavement distresses, the usability and adoption of automated pavement systems can be significantly improved. Finally, the ability to generate natural language descriptions alongside PCI values could lead to better-maintained infrastructure and more efficient resource use.

## LITERATURE REVIEW

### Deep Learning-Based PCI Estimation Methods

The Pavement Condition Index (PCI) is a critical metric for assessing road surface quality, essential for effective infrastructure management *(7–9)*. Traditional methods of PCI estimation, which rely on manual inspections, are often criticized for their subjectivity and inefficiency. To address these limitations, researchers have increasingly turned to deep learning techniques to automate PCI assessments and improve accuracy.

Early advancements in this field were made by Majidifard et al. *(3)* who proposed a hybrid model integrating YOLO for object detection with U-Net for image segmentation to classify and quantify pavement distresses. This approach provided a comprehensive assessment of pavement conditions but required large annotated datasets and significant computational power. These requirements, coupled with the challenges of ensuring high image quality, highlighted the need for further refinement in deep learning models for PCI estimation.

Following this, Han et al. (*6*) utilized aerial imagery combined with convolutional neural networks (CNNs) to evaluate pavement conditions. The model (PCIer) demonstrated high accuracy in categorizing pavement conditions across large areas. However, this approach faced challenges related to the need for specialized aerial equipment and the potential degradation of image quality due to environmental factors. Additionally, capturing fine-grained details essential for accurate PCI estimation proved difficult with aerial imagery.

In another significant development, Owor et al. (*10*) developed a multitask learning framework that combines detection, segmentation, and PCI estimation into a single model. This unified model enhanced the efficiency of PCI assessments while maintaining accuracy across tasks. However, the increased computational complexity and the potential for performance trade-offs between tasks posed significant challenges, limiting the framework's scalability.

Building on these efforts, Ibragimov et al. (*11*) introduced an automated system that uses deep learning for crack detection, followed by image processing algorithms for crack width estimation. While this method significantly improved the speed and accuracy of PCI assessments, it was heavily dependent on high-quality images and required substantial computational resources, limiting its applicability in varied environments.

Despite these advancements, current deep learning approaches for PCI estimation face several limitations, including dependency on high-quality data, substantial computational demands, and challenges in generalizing across diverse pavement conditions. These challenges underscore the necessity for developing a more streamlined and efficient PCI estimation framework that can operate effectively with diverse input data and reduced computational requirements. Such a framework would advance the field by addressing the current limitations and improving the scalability and practicality of deep learning-based PCI assessment tools.

### Image captioning

Image captioning is a process that generates a textual description for an image using Natural Language Processing (NLP) and Computer Vision techniques to automatically create descriptions of image content in natural language *(12, 13)*. Conventionally, image captioning relied on models like CNN-LSTM (*14*), which had limitations in capturing complex relationships between visual elements and textual descriptions. Biradar et al. (*15*) used a CNN-LSTM model for automatic image captioning, aiming to better capture the link between images and text. However, their model struggled with identifying key objects and their relationships in images, and generating grammatically and logically sound captions. Bindu et al. (*12*) proposed "captionify", a CNN-LSTM model which outperformed existing approaches in generating accurate captions by combining CNNs' visual feature extraction with LSTM's textual generation, enhancing accuracy and context relevance in image captioning. Lala et al. *(16)* developed an image captioning model using a modified Xception-based CNN and LSTM. While their model showed improved performance on datasets like Flickr8k, it struggled with understanding the semantics of complex scenes, potentially impacting caption quality for intricate images. To address the aforementioned limitations, Yan et al. *(29)* proposed a Task-Adaptive Attention module to improve image captioning by balancing visual and non-visual word generation. While showing promise, the module's complexity and potential for further optimization required further investigation. Despite these improvements, attention-based LSTMs still struggled with the inherent sequential processing constraints of RNNs, which limited their ability to parallelize computations and handle long-range dependencies efficiently *(30)*.

The introduction of Transformer architectures marked a significant leap forward in image captioning. Transformers, which rely on self-attention mechanisms, can process sequences in parallel, thereby overcoming the sequential bottlenecks of RNNs. This architecture allows for better handling of long-range dependencies and more efficient computation *(31, 32)*.

Yu et al. *(33)* proposed a Multimodal Transformer (MT) model for image captioning, utilizing multi-view visual features and a unified attention block to capture relationships within and between image and text data. Despite outperforming previous methods on the MSCOCO dataset, the model's complexity and computational cost remain limitations. Addressing the limitations of the MT model, Shao et al. *(34)* proposed a transformer-based dense image captioning architecture (TDC) that prioritizes more informative regions using a novel region-object correlation score unit (ROCSU). This method overcomes the vulnerability of LSTM in handling long sequences and the equal importance given to all regions in previous models. Despite its improvements, the TDC model still struggled with the computational overhead associated with the ROCSU. These studies suggest that Transformer architectures significantly improve image captioning performance by capturing complex interactions and prioritizing informative regions, although they often face challenges related to computational complexity.

In addition to the Transformer architecture, convolutional feedforward networks have also been integrated into image captioning models to enhance performance. These networks help in refining the visual features extracted by the CNN, ensuring that the most critical information is retained, and non-essential details are filtered out *(35)*. This combination of convolutional and Transformer-based approaches has been shown to outperform traditional methods significantly *(32)*.

**DATASET**

The dataset *(36)* used for this research consists of top-down views of pavement image data and its associated PCI only. The images were acquired under varying conditions from different sources in three different cities including: Jefferson City – Missouri, Peoria and Washington – Illinois. Figure 1 shows samples of these pavement images with their corresponding PCI displayed on the top. These images encompassed a diverse range of pavement conditions, including various types of distresses such as cracks, rutting, and potholes. The PCI for each image was calculated in accordance with ASTM standards and guidelines using crack type, extent, and severity of the pavement. However, no additional annotations or labels beyond the PCI values were provided in the original dataset. The PCI values associated with each image ranged from 0 to 100, where 0 represents the worst pavement condition, and 100 represents the optimal condition. Table 1 provides the ASTM-proposed standards for classifying PCI based on the distress type, severity, and extent. In addition to the PCI labels, we added bounding box annotations and textual annotations of the pavement condition. The following subsections provide details about these additional data annotation processes.

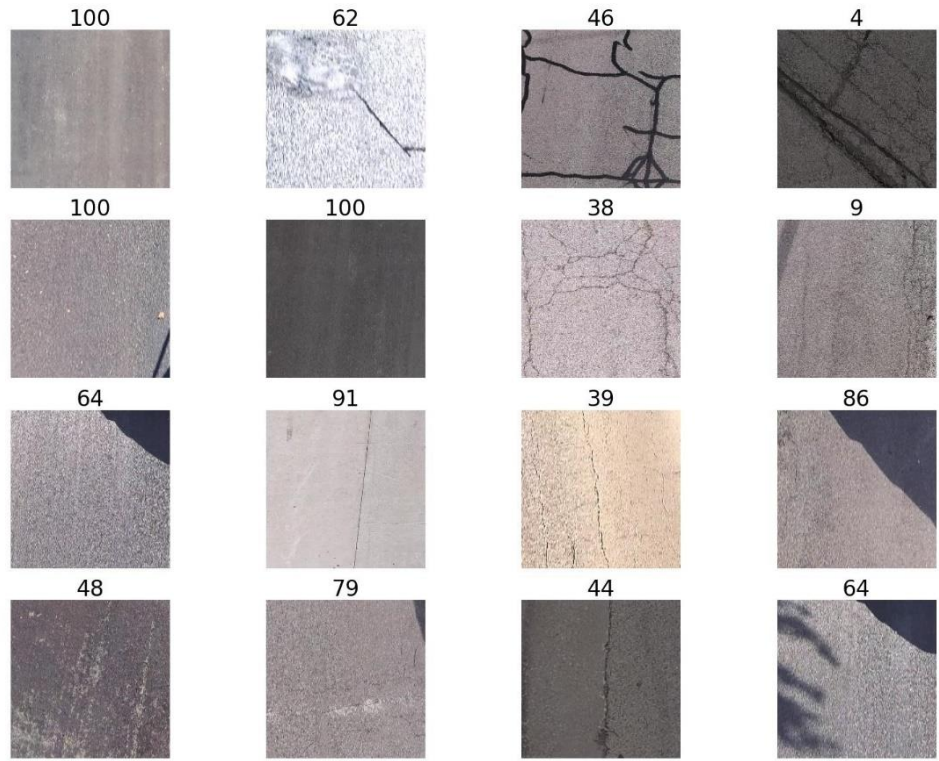

**Figure 1**. **Sample images from the dataset depicting various pavement conditions with their corresponding PCI displayed on the top**. The PCI labels seen in the figure range from 4 (severe distress) to 100 (optimal condition)

**Bounding Box Annotations**

To enhance the dataset for our multimodal approach, we conducted additional annotations on the pavement images using the Computer Vision Annotation Tool (CVAT) *(37)*. This process involved creating bounding boxes around specific distress type. Six primary distress types were considered for annotation: longitudinal cracking, transverse cracking, potholes, patching, block cracking, and diagonal cracking.

**Table 1**. **Classification of PCI Ranges**. This table presents the standard ASTM classification system for interpreting PCI values, which range from 0 to 100. It categorizes the PCI into distinct classes, such as 'Failed,' 'Serious,' 'Very Poor,' 'Poor,' 'Fair,' 'Satisfactory,' and 'Good'. The PCI is calculated after considering the distress type, extent, and severity on the pavement surface.

| PCI range | Class |
|---|---|
| 0-10 | Failed |
| 10-25 | Serious |
| 25-40 | Very Poor |
| 40-45 | Poor |
| 55-70 | Fair |
| 70-85 | Satisfactory |
| 85-100 | Good |

To ensure annotation quality and consistency, we developed a set of guidelines for the annotators, including minimum size thresholds for each distress type, rules for handling overlapping distresses, and criteria for distinguishing between similar distress types. Figure 2 shows samples of pavement images with various types of cracks and patching.

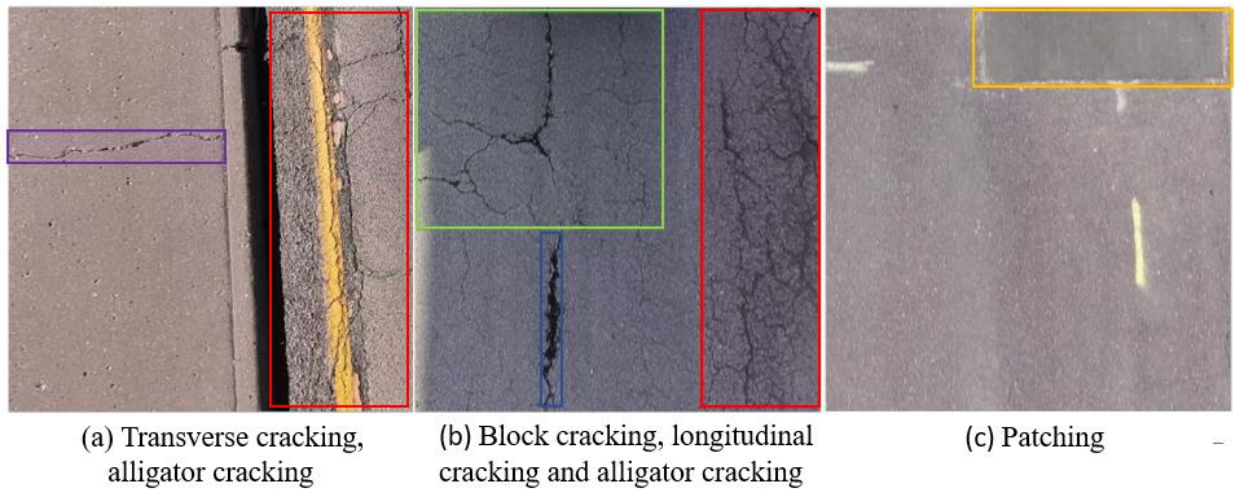

(a) Transverse cracking, alligator cracking  (b) Block cracking, longitudinal cracking and alligator cracking  (c) Patching

**Figure 2**. **Examples of pavement distress types with bounding box annotations**. (a) Transverse cracking (purple) and alligator cracking (red) along a road edge. (b) Multiple distress types including block cracking (green), longitudinal cracking (blue), and alligator cracking (red). (c) Patching (orange) on a section of pavement. These annotated images demonstrate the variety of distress types considered in our study and illustrate the bounding box annotation process used to enrich the dataset.

**Textual Annotations (Captions)**

To complement the bounding box annotations and PCI values, we developed a comprehensive textual annotation scheme for each pavement image. Figure 3 provides details about the textual annotation of a pavement image. This scheme was designed to provide a detailed and structured account of the pavement condition, significantly enhancing the dataset's richness and facilitating the training of our multimodal model. The textual annotations were meticulously crafted by pavement engineering experts to ensure accuracy and consistency across the dataset, following a standardized format comprising four key components.

The first component, pavement severity, offers an overall assessment of the pavement's condition, categorized as high, medium, or low severity based on the extent and nature of observed distresses. This is followed by a detailed list of present distresses, which includes specific types such as alligator cracking, longitudinal cracking, transverse cracking, potholes, and patching, along with their respective severity levels. To provide a comprehensive assessment, the third component explicitly mentions distress types not observed in the image, offering crucial negative information that aids in distinguishing between present and absent features. The final component incorporates the numerical value, providing a quantitative measure of the overall pavement condition.

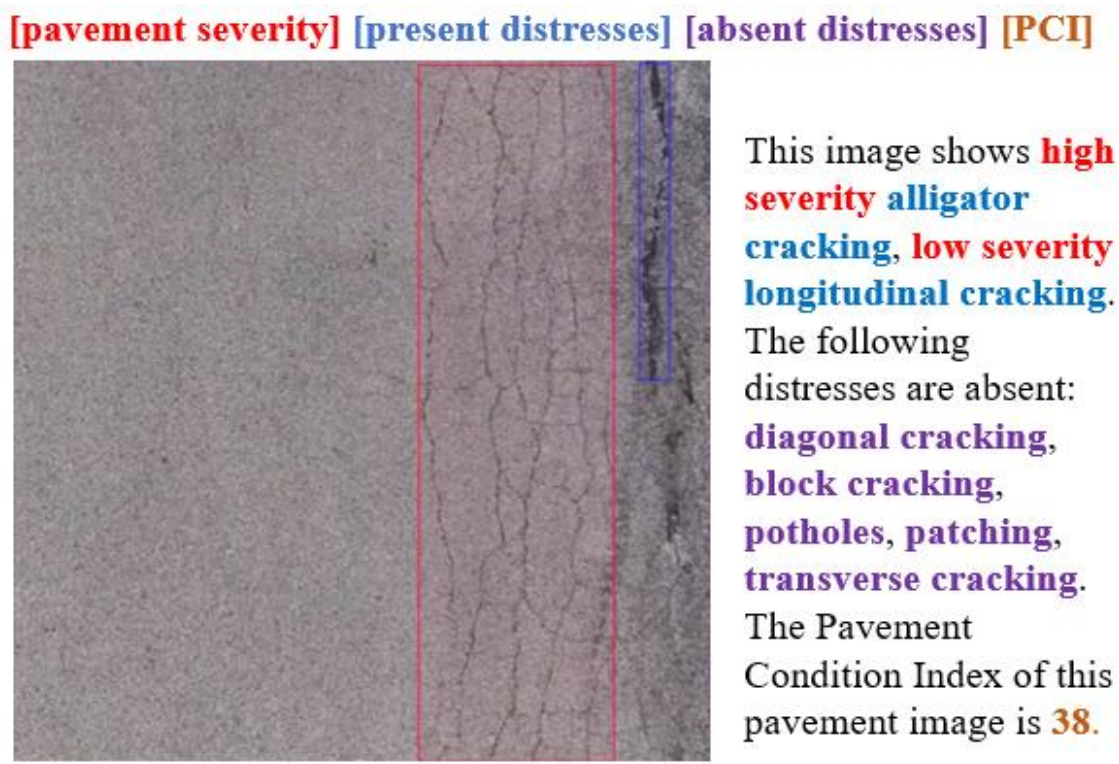

**Figure 3. An example of a pavement image and its corresponding caption in our manually annotated pavement dataset.** The caption includes different segments: pavement severity, present distresses, bounding box grounding, absent distresses, and the PCI. Each segment is indicated by a distinct color, shown at the top of the figure.

## PROPOSED METHODOLOGY

### Problem Formulation and Overview

Machine learning and deep learning approaches for pavement condition assessment has predominantly been treated as a unimodal problem *(38–51)* instead of a multi-modal problem, relying heavily on only image-based analysis for distress detection, segmentation, and classification. In this study, we formulate the problem as a multimodal task.

Let $I \in \mathbb{R}^{H \times W \times C}$ represent a pavement image with $H$ height, $W$ width, and $C$ channels. The goal is to estimate the PCI value from 0 to 100 and generate a dense caption that accurately describes the pavement condition. Our framework consists of two main components: a PCI prediction network and a dense captioning network. The PCI prediction network $f_{PCI}: \mathbb{R}^{H \times W \times C} \rightarrow [0, 100]$ is a neural network model with parameters $\theta$ designed to predict the PCI value from the input image $I$. The model $f_{PCI}$ can be expressed as $\hat{y} = f_{PCI}(I; \theta)$ where $\hat{y} \in [0,100]$ is the predicted PCI value. The dense captioning network $f_{CAP}: \mathbb{R}^{H \times W \times C} \rightarrow \mathcal{T}$ is a model with parameters $\emptyset$ that outputs a set of textual descriptions $\mathcal{T} = \{t_1, t_2, \ldots, t_n\}$, where $t_i$ represents individual textual descriptions. The model $f_{CAP}$ can be expressed as $\mathcal{T} = f_{CAP}(I; \theta)$. To generate the overall dense caption, the PCI prediction $\hat{y}$ and the textual descriptions $\mathcal{T}$ are combined. Let $f_{final}: [0,100] \times \mathcal{T} \rightarrow \mathcal{H}$ be a function with parameters $\psi$ that combines the PCI value and the set of textual descriptions to produce the overall dense caption $\mathcal{H}$. The overall dense caption $\mathcal{H}$ is represented as $\mathcal{H} = f_{final}(\hat{y}, \mathcal{T}; \psi)$.

Our overall framework involves a series of step. First, the pavement image is passed through a PCI Estimation network to predict the PCI. Simultaneously, the pavement image is passed through a dense captioning network to generate a dense caption. The results from both networks are combined to generate the final caption. Figure 4 illustrates the overall framework for our dense pavement captioning network. The next section goes into detail about the two main components of the framework.

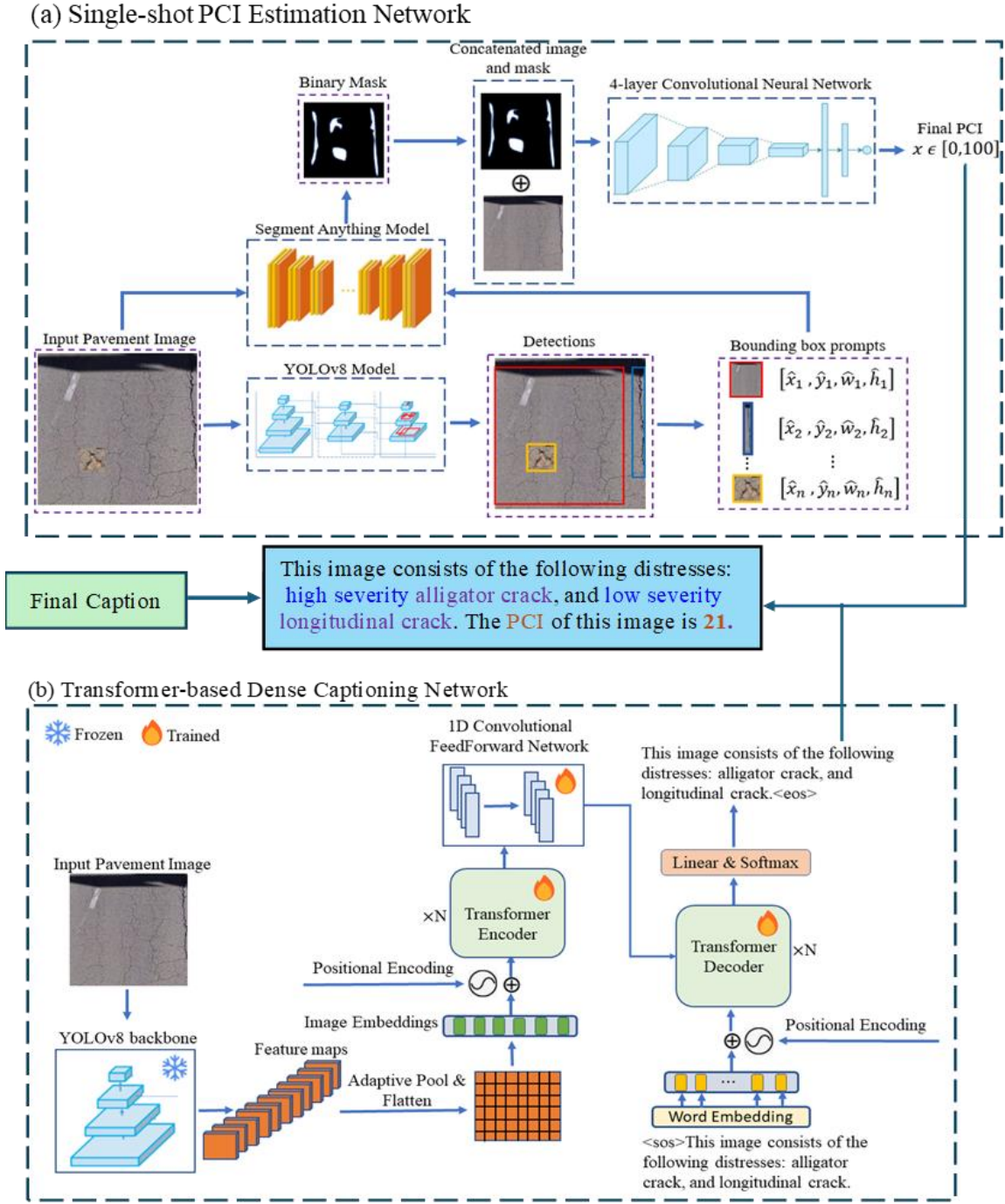

**Figure 4**. **Overall two-stage framework for our Dense Pavement Captioning Network**.

**Novel PCI Estimation Network**

We introduce a novel framework for PCI estimation, characterized by three core components: (1) a state-of-the-art object detection model, (2) a zero-shot object segmentation model, and (3) a four-layer convolutional neural network. The framework takes in pavement images processed through the YOLOv8 object detection model, which yields a set of bounding boxes and their associated classes. These bounding boxes, along with the input images, are subsequently passed to a zero-shot segmentation model (Segment Anything Model (SAM)), which outputs binary masks using the bounding boxes as prompts. The resulting binary mask is then concatenated with the input image and fed into the four-layer convolutional neural network, ultimately producing the final PCI value. The framework of the dense captioning network is illustrated in Figure 5. The three core components have been expanded below.

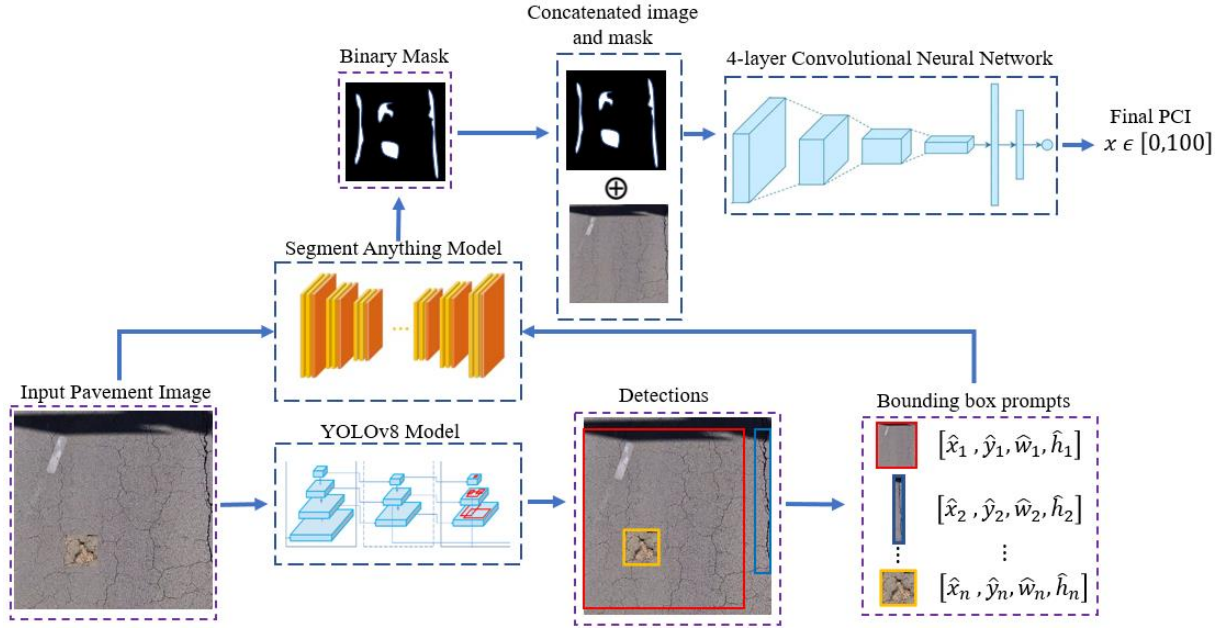

**Figure 5**. **PCI Estimation Network**.

*Accurate Object Detection Model (YOLOv8)*

In the initial phase of our framework, we utilize the YOLOv8 (You Only Look Once version 8) object detection model to identify and classify objects within pavement images. YOLOv8 is known for its efficiency and accuracy, performing object detection in a single forward pass. The model operates on an input image $I$ of dimensions $H \times W \times C$, where $H$ is the height, $W$ is the width and $C$ is the number of channels. The YOLOv8 model processes the image through a series of convolutional layers to extract hierarchical features, resulting in feature maps of varying spatial resolutions. The detection head of YOLOv8 predicts bounding boxes and class probabilities for each object detected in the image. The bounding box for object $i$ is predicted as:

$$\hat{B}_i = (\hat{x}_i, \hat{y}_i, \hat{w}_i, \hat{h}_i) \tag{1}$$

where $\hat{x}_i$ and $\hat{y}_i$ are the center coordinates, $\hat{w}_i$ and $\hat{h}_i$ are the width and height of the bounding box, respectively. The class probability for object is computed as:

$$\hat{P}(c_i|I) = \frac{exp(C_i \cdot f(I))}{\sum_j exp(C_i \cdot f(I))} \tag{2}$$

where $C_i$ is the weight vector for class $i$, and $f(I)$ represents the feature map extracted from the image. The overall loss function of YOLOv8 combines localization loss, confidence loss, and classification loss:

$$\mathcal{L}_{YOLO} = \lambda_{coord} \sum_{i=1}^{N} \mathcal{L}_{loc}\left(\hat{B}_i, B_i\right) + \lambda_{conf} \sum_{i=1}^{N} \mathcal{L}_{conf}\left(\hat{P}_{conf_i}, P_{conf_i}\right)$$
$$+ \lambda_{cls} \sum_{i=1}^{N} \mathcal{L}_{cls}\left(\hat{P}(c_i|I), P(c_i|I)\right)$$
(3)

where $N$ is the number of detected objects.

*Zero-shot Object Segmentation Model (SAM)*

Following object detection, the framework advances to the segmentation stage, employing SAM to generate precise binary masks for the detected objects. SAM leverages a zero-shot learning approach, meaning it can generate accurate segmentations without prior task-specific training. Given an input image $I$ of dimensions $H \times W \times C$ and bounding boxes $\hat{B}_i$ produced by YOLOv8, SAM generates binary masks $\hat{M}_i$ for each detected object.
The SAM model utilizes the bounding boxes as prompts to refine and produce binary segmentation masks:
$$\hat{M}_i = SAM(I, \hat{B}_i) \tag{4}$$

where $\hat{M}_i$ is the binary mask for object $i$ generated by SAM. The segmentation process involves feature extraction from the input image, integration of the bounding box prompts, and mask generation. The SAM model's loss function combines dice loss and binary cross-entropy loss:

$$\mathcal{L}_{SAM} = \lambda_{dice}\mathcal{L}_{dice}(\hat{M}_i, M_i) + \lambda_{bce}\mathcal{L}_{bce}(\hat{M}_i, M_i) \tag{5}$$

where $M_i$ is the ground truth mask for object $i$.

*Four-layer Convolutional Neural Network (CNN)*

The final stage of the framework involves a four-layer CNN designed to estimate the PCI value from the processed image data. The input to this network is a concatenation of the original pavement image and the binary masks generated by SAM, forming an input of dimensions $H \times W \times (C + 1)$. By concatenating the original pavement image with the binary masks, we create a composite input that retains both the raw visual details and the precise segmentation information. This concatenated input $I_{concat} = [I; \hat{M}_i]$ is fed into the CNN, which consists of four convolutional layers. Each layer extracts and processes features, ultimately producing the final PCI value. The operation of each convolutional layer can be described as:

$$O_i = \sigma(W_i * I_i + b_i) \tag{6}$$

where $O_i$ represents the output feature map of the $i$-th convolutional layer, $W_i$ and $b_i$ are the convolutional kernel weights and biases, $*$ denotes the convolution operation, and $\sigma$ is the activation function, typically ReLU. The final convolutional layer outputs a feature map $O_4$ that is flattened and passed through a fully connected layer to map the extracted features to the PCI value:

Agyei Kyem, Denteh, Asamoah and Aboah$$PCI = f_{CNN}(O_4) \tag{7}$$

where $f_{CNN}$ is the function representing the fully connected layer. The loss function for the CNN is the mean squared error (MSE) between the predicted $\widehat{PCI}$ and the ground truth $PCI$:

$$\mathcal{L}_{PCI} = \frac{1}{N}\sum_{i=1}^{N}(PCI_i - \widehat{PCI}_i)^2 \tag{8}$$

The final PCI value is added to the output of the textual descriptions generated by the dense captioning network. The next section talks about the components of the dense captioning network.

**Dense Captioning Network**

The study designed a dense captioning framework, composed of three main components: the YOLOv8 backbone network, a Transformer encoder-decoder architecture, and a convolutional feed-forward module. The process begins with comprehensive feature extraction through the YOLOv8 backbone network, trained on a large dataset of pavement images, resulting in detailed feature maps. These feature maps are then standardized via an adaptive pooling layer and reshaped into a one-dimensional feature vector. This vector is fed into an enhanced Transformer encoder-decoder architecture, where the encoder is augmented with a convolutional feed-forward layer consisting of two convolutional layers with a ReLU activation function, allowing for the extraction of intricate features. The encoded representations are concatenated with ground truth captions and processed through the Transformer's decoder blocks. The final output layer generates a probability distribution over the vocabulary for each token position, conditioned on the previously generated tokens. The framework of the dense captioning network is illustrated in Figure 6. The next section outlines the details of the main components of the network.

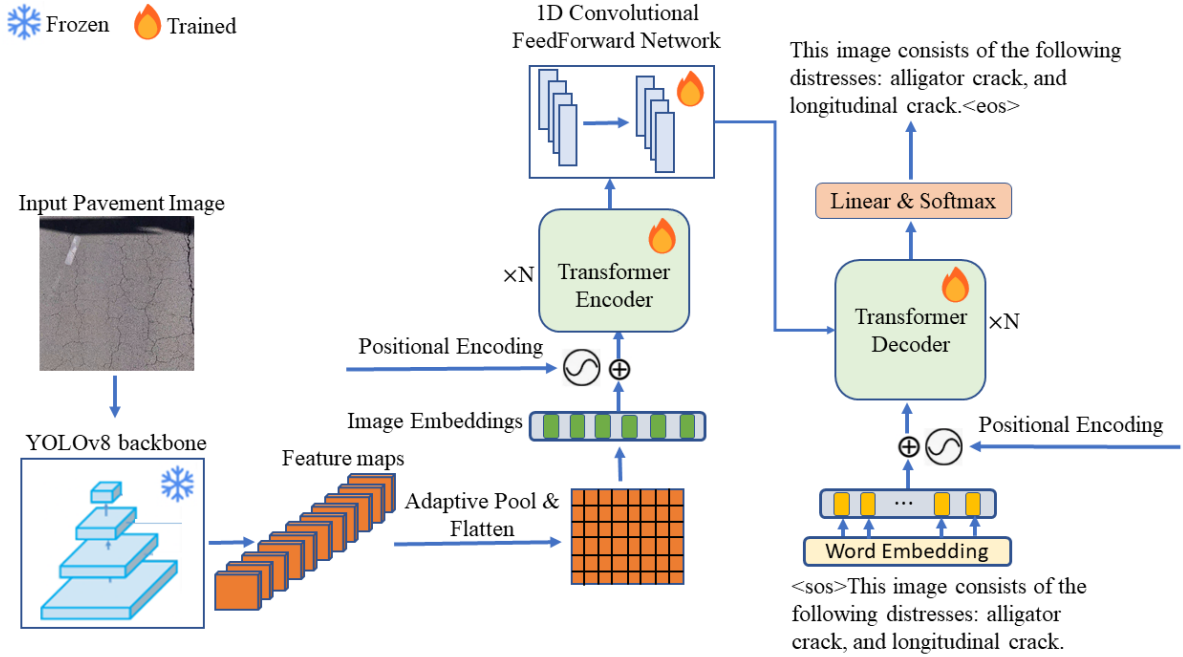

**Figure 6**. Dense Captioning Network

*YOLOv8 Backbone Network*

Our dense captioning framework involves utilizing the YOLOv8 backbone network for comprehensive feature extraction. YOLOv8, trained on a substantial dataset of pavement images, is adept at capturing essential visual details. Importantly, the backbone of YOLOv8 is frozen during training to retain its learned representations. The process starts with an input image $I$ of dimensions $H \times W \times C$ where $H$ is the height, $W$ is the width and $C$ represent the color channels. The YOLOv8 model processes this image to generate feature maps:

$$F = YOLOv8_{backbone}(I) \qquad (9)$$

These feature maps $F$, which encapsulate detailed visual information, are then standardized to a fixed size using an adaptive pooling layer. To ensure uniformity in subsequent processing steps, the feature maps $F$ are passed through an adaptive pooling layer. This layer resizes the feature maps to a predetermined size $H_P \times W_P$, making them suitable for further operations. The adaptive pooling operation is defined as:

$$F_{pooled}(i,j) = \max_{m,n} F\left(\left\lfloor \frac{m \cdot H}{H_p} \right\rfloor, \left\lfloor \frac{n \cdot W}{W_p} \right\rfloor\right) \qquad (10)$$

where $(i,j)$ are the corresponding coordinates in the original feature map, and $(m,n)$ adjusted according to the pooling window. Following the adaptive pooling layers, $F_{pooled}$ are flattened and reshaped into a one-dimensional feature vector. This transformation prepares the data for input into the Transformer encoder-decoder architecture. The process can be represented as:

$$f_{vector}(k) = F_{pooled}\left(\left\lfloor\frac{k}{W_p \cdot C}\right\rfloor, \left\lfloor\frac{k \bmod (W_p \cdot C)}{C}\right\rfloor, k \bmod C\right) \tag{11}$$

where $k$ ranges from 0 to $H_p \cdot W_p \cdot C - 1$.

*Transformer Encoder-Decoder Architecture*

The one-dimensional feature vector $\boldsymbol{f}_{vector}$ is then fed into a modified Transformer encoder-decoder architecture. This architecture is designed to handle sequential data and is enhanced by integrating a convolutional feed-forward layer within the encoder. The encoding process is expressed as:

$$E = Encoder(\boldsymbol{f}_{vector}) \tag{12}$$

Within the Transformer encoder, the self-attention mechanism allows the model to weigh the importance of different parts of the pavement input image sequence:

$$Attention(Q, K, V) = softmax\left(\frac{QK^T}{\sqrt{d_k}}\right)V \tag{13}$$

where $Q$ (queries), $K$ (Keys), and $V$ (Values) are the derived are derived from the input embeddings, and $d_k$ is the dimensionality of the keys.

Within the Transformer encoder, the convolutional feed-forward module plays a crucial role in refining the encoded features. This module consists of two convolutional layers with a ReLU activation function interspersed between them. The module enhances the model's ability to capture intricate and localized features. The operations are defined as:

$$E_{conv} = \sigma(W_1 * E + b_1) \tag{14}$$

$$E_{conv} = W_2 * E_{conv} + b_2 \tag{15}$$

where $\sigma$ denotes the ReLU activation function, $W_1$ and $W_2$ are the convolutional kernel weights, and $b_1$ and $b_2$ are the biases.

The encoded representations $E_{conv}$ are concatenated with the ground truth caption sequences $C_{gt}$ and passed into the Transformer decoder. The concatenation step ensures that the model has access to both visual and textual information:

$$D_{input} = [E_{conv}; C_{gt}] \tag{16}$$

The Transformer decoder processes these inputs sequentially to generate dense captions:

$$D = Decoder(D_{imput}) \tag{17}$$

The final stage involves generating a probability distribution over the vocabulary for each token position using a fully connected layer. This layer produces unnormalized logits, which are then normalized using the softmax function:

$$z_i = W_o \cdot D_i + b_o \tag{18}$$

$$P(t_i \mid t_{1:i-1}) = \frac{exp(z_i)}{\sum_{j=1}^{V} exp(z_j)} \tag{19}$$

where $P(t_i \mid t_{1:i-1})$ represents the probability of token $t_i$ given the sequence of previously generated tokens $t_{1:i-1}$.

The training of the model incorporates a combination of cross-entropy loss and doubly stochastic attention regularization. The cross-entropy loss $L_{CE}$ is given by:

$$L_{CE} = -\sum_{t=1}^{T} log(P(t_t \mid t_{1:t-1})) \tag{20}$$

where $T$ is length of the sequence

To encourage the model to distribute its attention across the entire image rather than focusing on specific parts, we add a doubly stochastic attention regularization term $L_{DSA}$:

$$L_{DSA} = \sum_{l=1}^{L} \frac{1}{L} \left( \sum_{d=1}^{D} \sum_{i=1}^{P^2} \left( 1 - \sum_{c=1}^{T} \alpha_{cidl} \right)^2 \right) \tag{21}$$

where $D$ is the number of heads, $L$ is the number of layers, $P$ is the size of the feature map, $\alpha_{cidl}$ is the attention weight.

The total loss then becomes:

$$L = -\sum_{t=1}^{T} log(P(t_t \mid \boldsymbol{t}_{1:t-1})) + \sum_{l=1}^{L} \frac{1}{L} \left( \sum_{d=1}^{D} \sum_{i=1}^{P^2} \left( 1 - \sum_{c=1}^{T} \alpha_{cidl} \right)^2 \right) \tag{22}$$

**Evaluation Metrics**

*PCI Estimation Network*

In the evaluation of the PCI Estimation Network, we employed two primary metrics: Mean Squared Error (MSE) and Mean Absolute Error (MAE). These metrics were chosen for their effectiveness in quantifying the accuracy of continuous predictions, particularly in the context of regression tasks like pavement condition index estimation.

Mean Squared Error (MSE)

MSE is a widely used metric that measures the average of the squares of the errors—that is, the average squared difference between the estimated values and the actual value.

$$MSE = \frac{1}{n} \sum_{i=1}^{n} (\hat{y}_i - y_i)^2 \tag{23}$$

where $n$ is the number of data points, $\hat{y}$ is the predicted PCI value for the $i$-th data point, and $y_i$ is the actual PCI value for the $i$-th data point.

Mean Absolute Error (MAE)
MAE measures the average of the absolute differences between the predicted values and the actual values. The MAE is computed as follows:

$$MAE = \frac{1}{n}\sum_{i=1}^{n}|\hat{y}_i - y_i|$$

(24)

*Dense Captioning Network*
For evaluating the Dense Captioning Network, we utilized three well-established metrics: BLEU (Bilingual Evaluation Understudy), METEOR (Metric for Evaluation of Translation with Explicit ORdering), and GLEU (Google's BLEU). These metrics are widely recognized for their ability to measure the quality of generated text against reference texts, making them particularly suitable for assessing the performance of captioning models.

BLEU
BLEU is a precision-based metric that evaluates the extent to which n-grams in the generated text match the n-grams in the reference text. This metric is particularly effective for machine translation and text generation tasks. The calculation of the BLEU score involves several steps. First, the n-gram precision is computed as follows:

$$P_n = \frac{\sum_{count} min\left(count_{clip}(n-gram), count_{ref}(n-gram)\right)}{\sum_{count} count(n-gram)}$$

(25)

where $P_n$ is the precision for n-grams, $count_{clip}$ is the clipped count of matching n-grams, and $count_{ref}$ is the count of n-grams in the reference text. The brevity penalty (BP) is then calculated to account for the length of the generated text relative to the reference text:

$$f(x) = \begin{cases} 1 & if\ c > r \\ e^{(1-r/c)} & if\ c \leq r \end{cases}$$

(26)

where $c$ is the length of the candidate text, and $r$ is the length of the reference text. The final BLEU score is computed as:

$$BLEU = BP \cdot exp\left(\sum_{n=1}^{N} w_n \log P_n\right)$$

(27)

where $w_n$ is the weight assigned to the n-gram precision, typically uniform.

METEOR
METEOR is a recall-based metric that aligns words in the generated text with words in the reference text using synonyms, stemming, and paraphrasing. It considers precision, recall, and an alignment score, making it more robust in capturing linguistic variations. The METEOR score is calculated as follows:

$$METEOR = F_{mean} \cdot (1 - P_{frag}) \tag{28}$$

where

$$P_{frag} = 0.5 \left(\frac{chunks}{u_m}\right)^3, \tag{29}$$

$$F_{mean} = \frac{10PR}{R + 9P}, \tag{30}$$

$$P = \frac{u_m}{c_m}, \tag{31}$$

and

$$R = \frac{u_m}{r_m} \tag{32}$$

where $chunks$ is the number of matching chunks (a chunk is a consecutive sequence of matched words in both the candidate text and the reference text), $u_m$ is the number of mapped words between the candidate text and the reference text, $c_m$ $and$ $r_m$ are the word counts in the candidate and the reference text respectively.

GLEU

GLEU, a variant of BLEU, addresses some of its shortcomings by considering both precision and recall of n-grams. It is particularly used for evaluating machine translation. The GLEU score is calculated by counting the matching n-grams between the generated and reference texts:

$$matches = \sum_{n-grams} min\left(count_{candidate}(n - gram), count_{reference}(n - gram)\right) \tag{33}$$

Next, both the precision and recall are computed:

$$precision = \frac{matches}{\sum_{n-grams} count_{candidate}(n - gram)} \tag{34}$$

$$recall = \frac{matches}{\sum_{n-grams} count_{reference}(n - gram)} \tag{35}$$

GLEU score is computed as:
$$GLEU = min(precision, recall) \tag{36}$$

**RESULTS**

**PCI Estimation Network**

*Qualitative results of the PCI network*

Our single-shot PCI estimation network demonstrated exceptional performance in predicting the exact PCI values for some of the images. This exceptonal performance can be attributed to several key factors. Firstly, the high precision of the YOLOv8 object detection accurately localizes pavement defects, such as cracks and potholes, ensuring that relevant features are captured effectively. The detailed segmentation masks generated by the SAM model further enhance this accuracy by providing precise boundaries of the defects. Additionally, the concatenation of the pavement images with these binary masks creates enriched feature representations, which the four-layer convolutional neural network (CNN) leverages to learn robust and discriminative features. This dual representation of visual and segmented data significantly enhances the model's ability to generalize across diverse pavement conditions. The results have been illustrated in Figure 7.

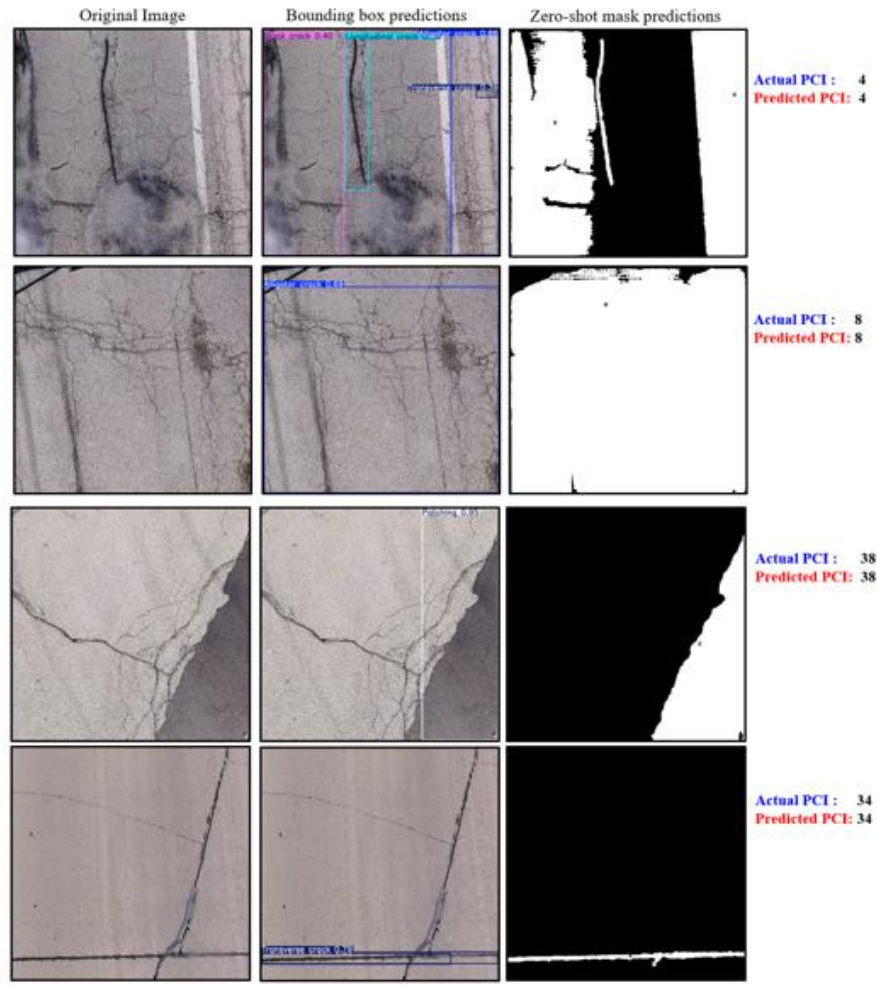

**Figure 7**. **Exact PCI predictions for samples of images**.

While our network generally performs exceptionally well, there are instances of good but not perfect predictions. For example, some predictions were slightly off, either marginally higher or lower than the actual PCI values, as shown in Figure 8. These minor discrepancies can be attributed to subtle defect patterns that are challenging to detect and segment accurately, variations in lighting and shadowing conditions, fixed road elements such as drains, and the complexity of mixed defect types in some pavements. Additionally, the object detection model occasionally failed to recognize these challenging distresses, considering some of the images as background and generating entirely null masks. Despite these challenges, the model's predictions remain close to the actual PCI values, demonstrating its robustness.

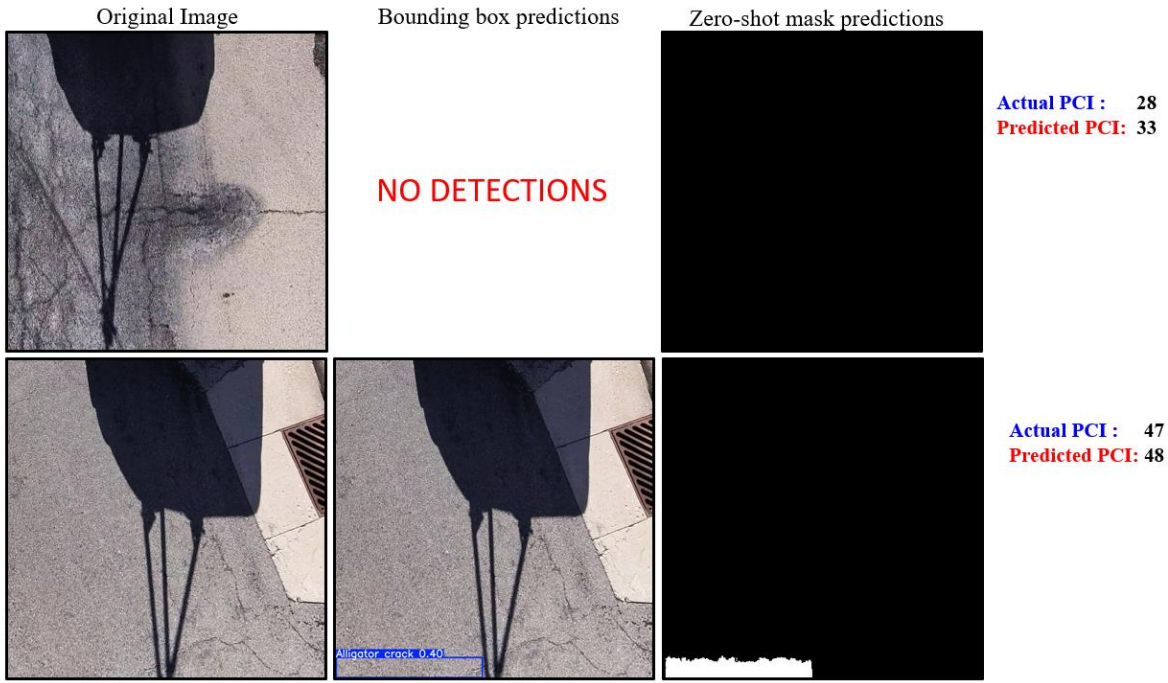

**Figure 8**. **Instances of slightly Off PCI Predictions**. The first row shows an example where the network produced null mask due to no detections by the object detection model. The second row displays a prediction affected by static road elements such as drains, leading to marginally higher or lower PCI values.

In some instances, our network exhibited poor performance, highlighting areas for potential improvement as illustrated in Figure 9. A significant factor contributing to this is the quality of the segmentation masks generated by the SAM model. Some masks fail to accurately delineate the patterns in complex distresses such as alligator cracks and block cracks often resulting in masks that cover the entire area of the bounding box rather than the specific distress. Additionally, the network occasionally produces incorrect detections, leading to erroneous masks that do not correspond to actual pavement defects. There are also cases where the model fails to detect all the distresses present in the image, or misses detecting any distresses altogether, treating parts of the pavement as background. These issues in the detection and segmentation stages propagate through the network, affecting the subsequent feature extraction and learning processes of the CNN. When the input to the CNN includes inaccurately segmented masks or incorrect detections, the resulting feature representations are flawed, leading to significant deviations in the predicted PCI values.

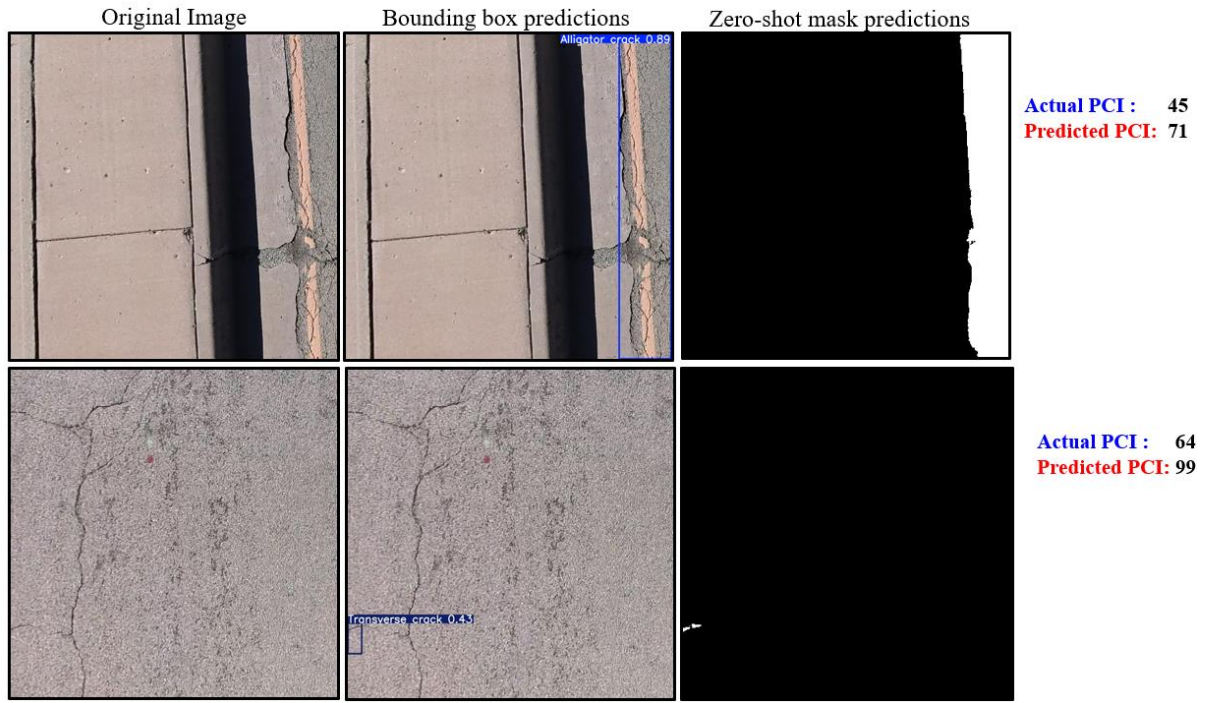

**Figure 9**. **Instances of Poor Performance in PCI Estimation**. The first row illustrates a scenario where the SAM model fails to delineate some of the detected cracks accurately, resulting in segmentation masks that incorrectly cover the entire bounding box area. The second row shows an example where the detection model fails to identify some of the present pavement distresses, leading to the SAM model generating null masks.

*Error Analysis of the Network*

To thoroughly evaluate the performance of our network, we conducted an error analysis focusing on the overprediction and underprediction of PCI values. Our analysis revealed that, on average, when the predicted PCI is higher than the actual PCI, the mean absolute error is 37.79 with a standard deviation of 24.28. Also, when the predicted PCI is lower than the actual PCI, the mean absolute error is 30.24 with a standard deviation of 25.24. Figure 10 illustrates the distributions of these errors. Despite these errors, the model's predictions generally remain within reasonable PCI severity ranges, as detailed in Table 1. This further illustrate that while the model's predictions may not always be exact, they still do not change the actual classification of the pavement condition. This robustness is crucial for practical applications where exact PCI values are less critical than the overall condition classification.

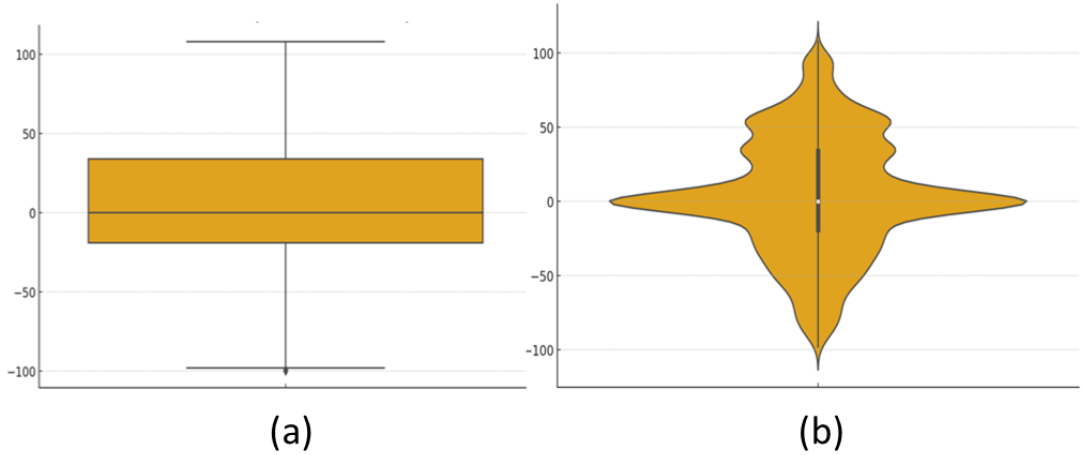

(a)                      (b)

**Figure 10. Error Distribution in PCI Estimation Network**. The boxplot (left) and violin plot (right) illustrate the distribution of prediction errors in the network. (a) A boxplot showing the interquartile range, median, and outliers of the errors. (b) A violin plot providing a density estimate for the distribution and frequency of the prediction errors. The majority of prediction errors are centered around zero, indicating that the model's predictions are generally close to the actual PCI values.

To understand the frequency of errors and identify the PCI ranges where the model tends to overpredict and underpredict, we plot a frequency distribution and a heatmap of the prediction errors as illustrated in Figure 11.

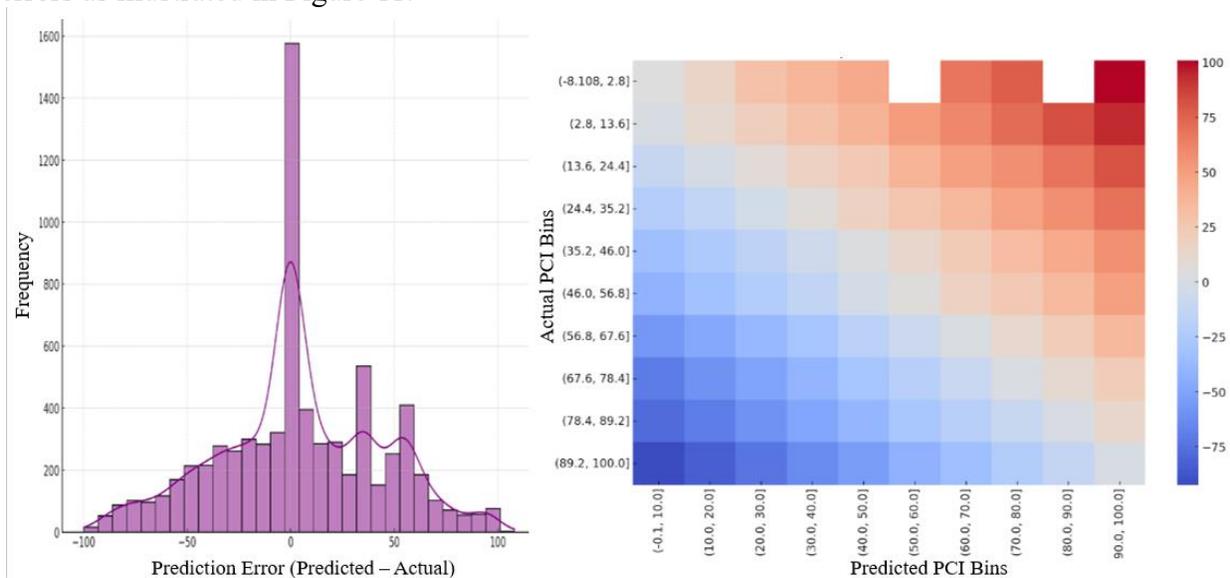

**Figure 11**. **Frequency Distribution and Heatmap of Prediction Errors.** The left plot shows the frequency distribution of prediction errors. The right heatmap visualizes mean prediction errors across PCI ranges, indicating overprediction (red) and underprediction (blue) areas.

To further analyze the network's performance, we plotted the predicted PCI values against the actual PCI values (Figure 12). This scatter plot includes a regression line to illustrate the relationship between the predicted and actual values, with a correlation value of approximately 0.70, indicating a strong positive correlation.

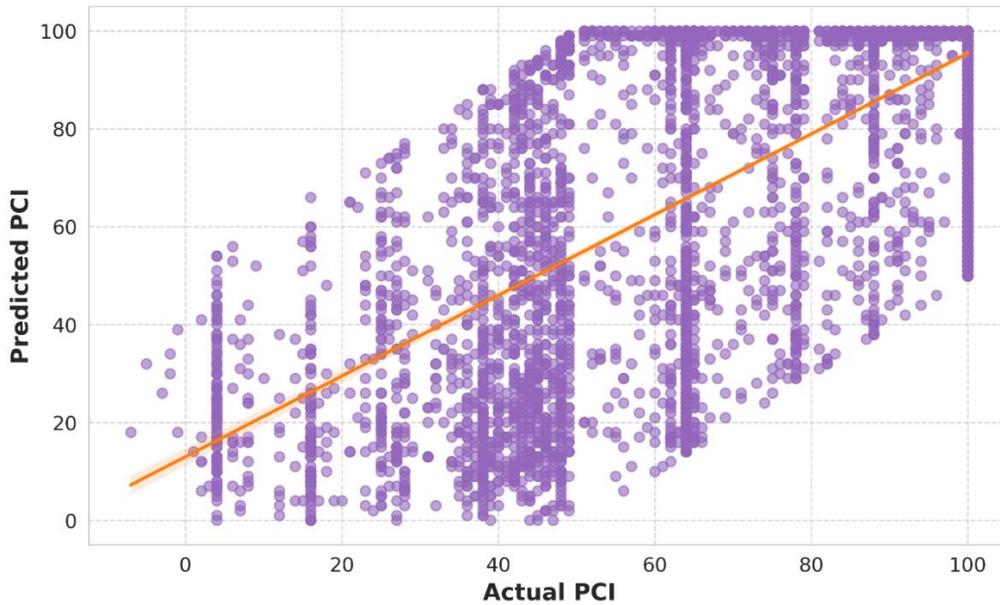

**Figure 12**. **Scatter Plot of Predicted PCI vs Actual PCI**

**Dense Captioning Network**

Our dense captioning network equally demonstrated outstanding performance in generating detailed and accurate captions for pavement images. This performance is as a result of the unique components of our dense captioning network such as the YOLOv8 backbone and the convolutional feedforward module. The YOLOv8 backbone, trained specifically on pavement images, effectively extracts detailed hierarchical features, capturing both low-level textures and high-level semantic information relevant to pavement distresses. These rich feature maps serve as a robust input for the Transformer encoder-decoder, which uses attention mechanisms to focus on relevant image areas. The convolutional feed-forward layers further refine these features, enabling the model to generate accurate and contextually precise dense captions. This integration allows the model to reliably describe various pavement conditions, as illustrated by the alignment of predicted captions with actual descriptions in Figure 13.

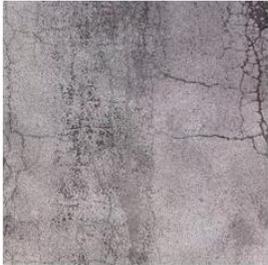
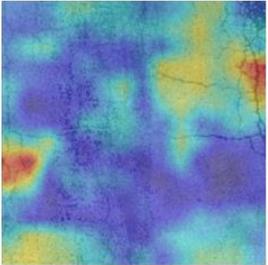
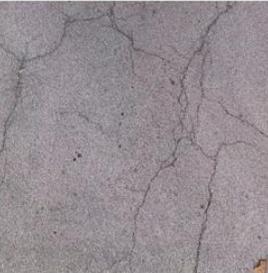
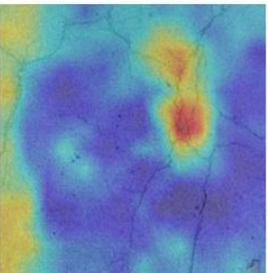
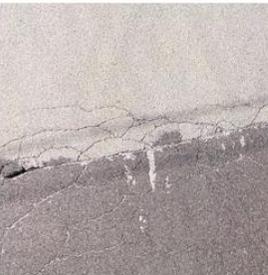
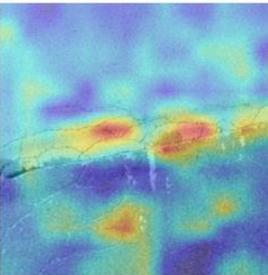

**Figure 13. Examples of predicted accurate captions for sample pavement images**. The figure shows the original images, attention maps, and captions. The attention maps highlight areas of focus of the model in the pavement image. The predicted captions closely match the actual descriptions, accurately identifying the types and severity of the distresses and noting the absence of other distresses.

However, there are instances where the network failed to generate accurate captions for the pavement images. For instance, in Figure 14 (a), the network failed to include 'low severity patching' as part of the distresses present in the image. This omission may be due to the model's reduced sensitivity to less prominent features, likely stemming from an imbalance in the training dataset where such features are underrepresented. In Figure 14 (b), the network completely failed to identify 'medium severity longitudinal cracking,' which suggests limitations in the model's ability to generalize to certain distress types.

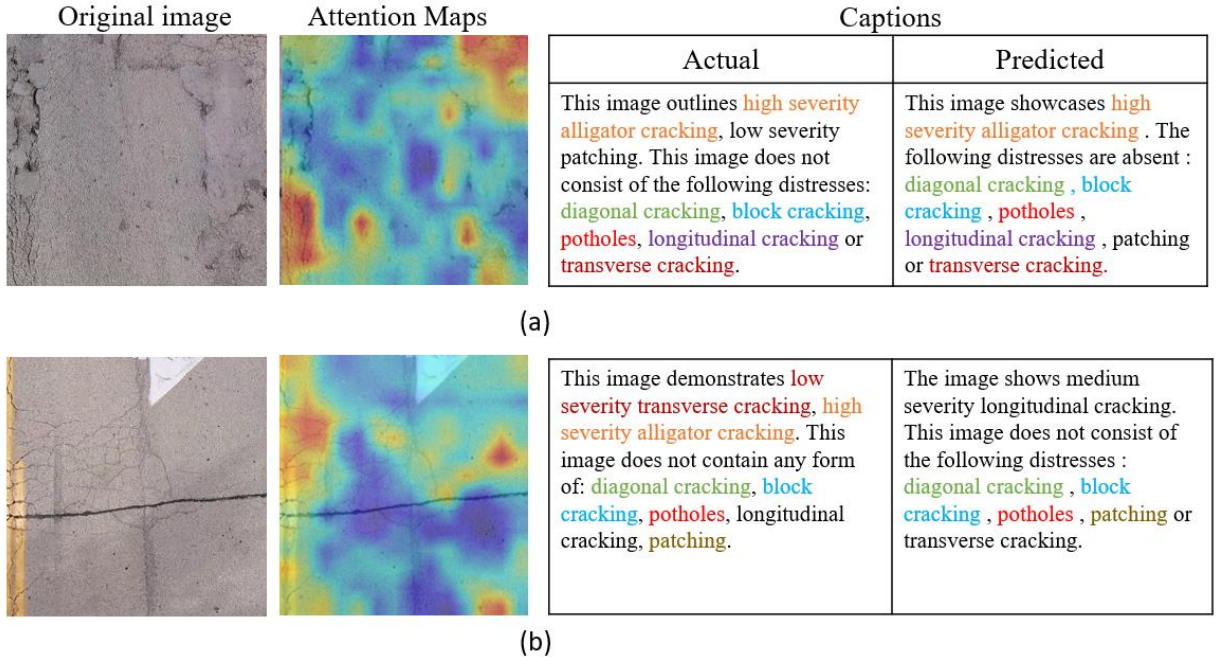

**Figure 14**. **Instances of the Dense Captioning Network's shortcomings**. (a) The network misses 'low severity patching' in the caption. (b) The network fails to identify 'medium severity longitudinal cracking.' These examples highlight areas for improvement.

An exceptional scenario is illustrated in Figure 15, where our network demonstrated its ability to correct an annotation error in the ground truth. The original ground truth annotation failed to include alligator cracking as part of the present distresses, while the model accurately generated a caption highlighting high severity alligator cracking, which was the correct distress type. This instance underscores the model's robustness to accurately identify pavement distresses even when the provided annotations are erroneous. This capability is particularly valuable in real-world applications, where annotation errors can be common, and the model's ability to self-correct contributes significantly to the overall accuracy and reliability of the inspection process.

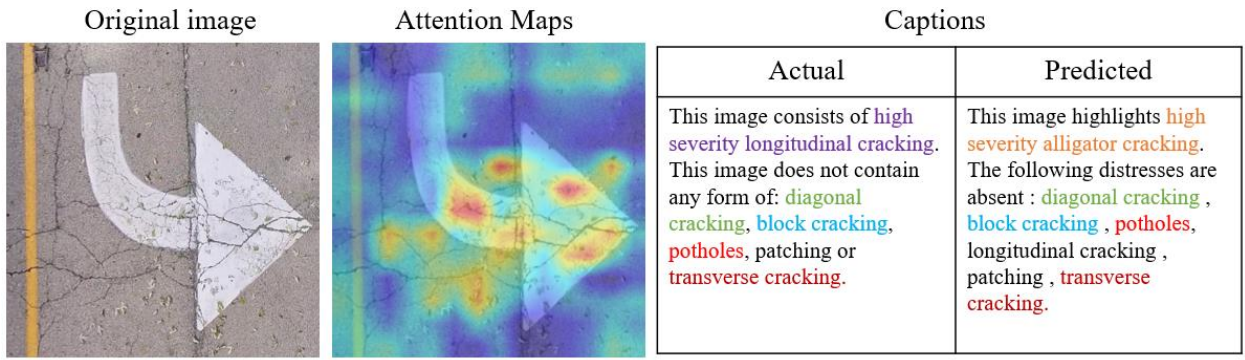

**Figure 15**. **Model correcting annotation error**. The original ground truth annotation identifies 'high severity longitudinal cracking,' while the model accurately generates 'high severity alligator cracking.' This instance highlights the model's ability to correct annotation errors.

Figure 16 illustrates the overall framework for our novel dense captioning network.

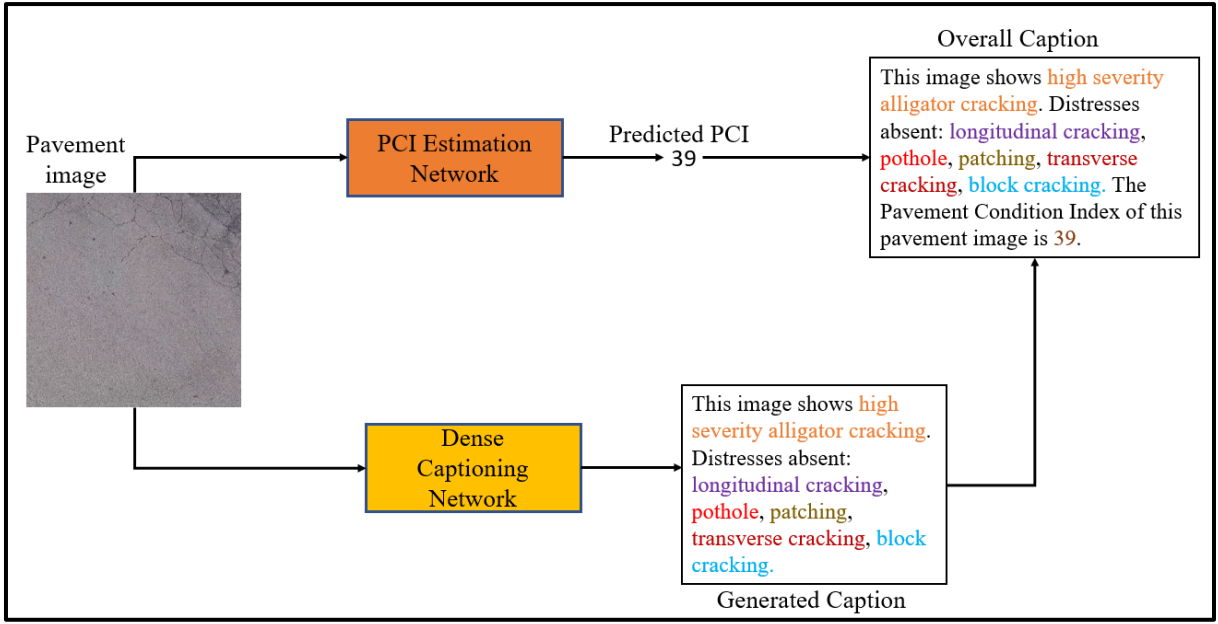

**Figure 16. Overall Framework for the Dense Captioning Network** (Inference Pipeline). The framework processes a sample pavement image through both the PCI Estimation Network and the Dense Captioning Network. The PCI Estimation Network predicts a PCI of 39, while the Dense Captioning Network generates a detailed caption identifying the distress types. The final output combines these results into an overall caption, providing both the predicted PCI value and the identified distresses for a comprehensive assessment of the pavement condition.

We evaluated the performance of our dense captioning model on our test data using several standard metrics, including BLEU (1 to 4 n-grams), GLEU, and METEOR. The following table presents the mean and standard deviation for each metric based on our test results.

**Table 2**. **Mean and standard deviation of evaluation metrics for dense captioning**. The mean values indicate the average performance, while the standard deviations reflect the variability in the model's outputs across different captions.

| Metric | Mean | Standard deviation |
|---|---|---|
| BLEU-1 | 0.7445 | ±0.195 |
| BLEU-2 | 0.6766 | ±0.212 |
| BLEU-3 | 0.6130 | ±0.238 |
| BLEU-4 | 0.5690 | ±0.261 |
| GLEU | 0.5893 | ±0.241 |
| METEOR | 0.7252 | ±0.199 |

**CONCLUSION**

This study addressed the challenge of automating pavement condition assessment by developing a novel multimodal framework combining computer vision and natural language processing. Our approach consisted of a PCI Estimation Network and a Dense Captioning Network, integrating visual information with natural language descriptions.

The PCI Estimation Network achieved a strong positive correlation (0.70) between predicted and actual PCIs, demonstrating its effectiveness in automating condition assessment. The Dense Captioning Network generated accurate pavement condition descriptions, as evidenced by high BLEU-1 (0.7445), GLEU (0.5893), and METEOR (0.7252) scores. Notably, the model showed robustness in handling complex scenarios, even correcting some annotation errors in the ground truth data.

For future work, we recommend exploring advanced vision-language models for pavement assessment tasks. This exploration necessitates the creation of a diverse multi-modal dataset combining pavement images and corresponding textual descriptions. With this enhanced dataset, we aim to expand beyond image captioning to tackle more complex tasks such as Visual Question Answering, Visual Reasoning, and Visual Grounding. By incorporating these advanced capabilities, we anticipate significant improvements in the interpretability and versatility of automated pavement assessment systems. Ultimately, these advancements have the potential to revolutionize infrastructure management and decision-making processes in the field of pavement maintenance.

**AUTHOR CONTRIBUTIONS**

The authors confirm contribution to the paper as follows: study conception and design: Blessing Agyei Kyem, Armstrong Aboah; data preparation: Eugene Kofi Okrah Denteh, Blessing Agyei Kyem, Joshua Kofi Asamoah; analysis and interpretation of results: Blessing Agyei Kyem; draft manuscript preparation: Blessing Agyei Kyem, Armstrong Aboah. All authors reviewed the results and approved the final version of the manuscript.